\documentclass[conference]{IEEEtran}

\usepackage{amsmath}
\usepackage{amssymb}
\usepackage{bm}
\usepackage{graphicx}
\usepackage{subcaption}
\usepackage[skip=0pt,font=small,labelfont=bf]{caption}
\usepackage{wrapfig}
\usepackage{adjustbox}
\usepackage{pgfplots}
\usetikzlibrary{patterns}
\usepackage{color,soul}
\usepackage{amssymb}
\usepackage{url}
\usepackage{breqn}
\usepackage{multirow}
\usepackage[top=72pt, left=54pt, right=54pt, bottom=54pt]{geometry}
\usepackage{cite} 
\usepackage[ruled,linesnumbered]{algorithm2e}
\usepackage[noend]{algpseudocode}

\usepackage{fixme}
\fxsetup{status=draft, theme=color}
\definecolor{fxtarget}{rgb}{0.8000,0.0000,0.0000}
\definecolor{fxnote}{rgb}{0.0000,0.0000,0.8000}

\IEEEoverridecommandlockouts                              

\title{\LARGE \bf
	Multi-Fingered Active Grasp Learning
}

\author{Qingkai Lu$^{1}$, Mark Van der Merwe$^{1}$, and Tucker Hermans$^{1,2}$%

\thanks{$^{1}$Qingkai Lu, Mark Van der Merwe, and Tucker Hermans are
  with the School of Computing and the Robotics Center, University of
  Utah, Salt Lake City, UT USA. $^{2}$NVIDIA; Seattle, WA, USA.
 {\tt\footnotesize qklu@cs.utah.edu; mark.vandermerwe@utah.edu;
   thermans@cs.utah.edu}. Q.~Lu was supported by NSF Award \#1846341
 and DARPA Award under grant N66001-19-2-4035. M.~Van der Merwe was supported in part by the Undergraduate Research Opportunities Program at the University of Utah.}%
}

\begin{document}

\maketitle

\begin{abstract}
Learning-based approaches to grasp planning are preferred over
analytical methods due to their ability to better generalize to new,
partially observed objects. However, data collection remains one of
the biggest bottlenecks for grasp learning methods, particularly for
multi-fingered hands. The relatively high dimensional configuration
space of the hands coupled with the diversity of objects common in
daily life requires a significant number of samples to produce robust
and confident grasp success classifiers. In this paper, we present the
first active deep learning approach to grasping that searches over the
grasp configuration space and classifier confidence in a unified
manner.
We base our approach on recent success in planning multi-fingered grasps as
probabilistic inference with a learned neural network likelihood function.
We embed this within a multi-armed bandit formulation of sample
selection.
We show that our active grasp learning approach uses fewer training samples to produce grasp success rates comparable with the passive supervised learning method trained with grasping data generated by an analytical planner.
We additionally show that grasps generated by the active learner have
greater qualitative and quantitative diversity in shape.
\end{abstract}



\section{Introduction}
\label{sec:intro}
Learning-based grasp planning~\cite{bohg2014data, saxena2006learning, Saxena-aaai2008, lenz2015deep, pinto2016supersizing, Kopicki2016, mahler2017dex, lu2017grasp,mousavian20196} has become popular over the past decade, because of its ability to generalize well to novel objects with only partial-view object information~\cite{sahbani2012overview}.
These approaches require large amounts of data for training, particularly those that utilize deep neural networks~\cite{varley2015generating, veres2017modeling, lu2017grasp, lu2020multi,mousavian20196}.
However, large scale data collection remains a challenge for
multi-fingered grasping, because (1) objects common in daily life
exhibit large variation in terms of geometry, texture, inertial
properties, and appearance; and (2) the relatively high dimension of
multi-fingered grasp configurations, (e.g. \(22\) dimensions for the
configuration of hand and wrist pose in this paper).

Random sampling is a common approach for data collection of two-finger
gripper grasping~\cite{mahler2017dex, pinto2016supersizing,
  levine2016learning}. Considering the sparsity of successful grasps
in the high dimensional configuration space of multi-fingered
grasping, it is not practical to collect enough successful grasps
using random sampling alone. Eppner et al.~\cite{eppner2019billion}
exhaustively sample more than 1 billion grasps for each of 21 objects
from the YCB data set to improve the understanding of data generation
for 6-DOF, parallel jaw grasp learning algorithms. However, this
approach would not scale well to multi-fingered grasping due to its
high dimensionality.

Existing deep learning work~\cite{varley2015generating,
  veres2017modeling, lu2017grasp} for multi-fingered grasping collect
training grasps using analytical or heuristic grasp planners to combat
the issues of sampling in high-dimensional spaces.
While these passive supervised grasp learning techniques generalize well over the
space covered by the training data, the grasp planners used for training bias data
collection to only cover a subspace of all feasible grasps configurations.
Additionally grasps generated for different object geometries tend to
be quite similar in shape.

We propose an active learning approach to interactively learn a grasp
model that better covers the grasp configuration space across
different objects using fewer samples compared with a passive, supervised grasp learner.
Instead of passively inducing a hypothesis to explain the available training data as in standard supervised learning, active learning develops and tests new hypotheses continuously and interactively.
Active learning is most appropriate when 1) unlabeled data samples are numerous, 2) a lot of labeled data are needed to train an accurate supervised learning system, and 3) data samples can be easily collected or synthesized~\cite{settles2012active}.
Grasp learning satisfies each of these conditions: 1) there are
infinitely many possible grasps, 2) a large number of labeled training
samples are necessary to cover the space~\cite{eppner2019billion}, and
3) the robot is its own oracle---it can try a grasps and automatically detecting success or failure without human labeling.

We propose modeling active grasp learning as a multi-armed bandit problem, which is designed to improve the grasp success classifier and enable the grasp model to cover the space of grasp configurations and objects as much as possible.
Our proposed method is fundamentally different from existing
bandit-based grasping work~\cite{kroemer2010combining,
  laskey2015multi, mahler2016dex}, which treat each possible grasp as
one arm and perform grasp planning separately for different
objects. We instead use the bandit-framework to select between
three qualitatively different exploration strategies each implemented
using the grasp planning as probabilistic inference formulation~\cite{lu2020multi}.

We perform grasp planning using our actively learned grasp model to
generate high-quality grasps for novel objects with different shapes
and textures. Our work demonstrates the first deep, active learning
approach to robotic grasping. Our real-robot grasping experiments show
our active grasp planner achieves comparable success rates, while
using fewer training data, when compared to a passive supervised
planner~\cite{lu2017grasp, lu2019grasp, lu2020multi} trained on data
generated with a geometric grasp planner. Furthermore our
active learner generates grasps with greater diversity than the passively trained model.

In the next section we review the literature of grasp planning and
active learning in robotic grasping. In Section~\ref{sec:grasp_inf} we
introduce our grasp planning as inference framework and define our
grasp model. We then describe our novel, bandit-based active grasp
learning algorithm in Section~\ref{sec:active_learning}. In
Section~\ref{sec:experiments}, we give a thorough account of our
experimental evaluation. In Section~\ref{sec:discussion}, we conclude
with a brief discussion and suggest directions for future work.


\section{Related Work}
\label{sec:related_work}

Broadly speaking, supervised deep grasp learning methods either predict grasp success from an image patch associated with a grasp configuration~\cite{lenz2015deep,gualtieri2016high,pinto2016supersizing,levine2016learning,mahler2017dex,johns2016deep,varley2015generating,mousavian20196} or directly predict a grasp configuration from an image or image patch using regression~\cite{redmon2015real, kumra2016robotic, veres2017modeling}. 
Most deep grasp learning approaches such as~\cite{bohg2014data, lenz2015deep, pinto2016supersizing, Kopicki2016, mahler2017dex, mousavian20196, kalashnikov2018qt} focus on parallel jaw grippers, while relatively little work has focused on the more difficult multi-fingered grasping problem~\cite{lu2017grasp, lu2020multi, varley2015generating, veres2017modeling, wu2020generative}.
Deep learning requires large amount of training data, however, it is
challenging to collect a large amount of grasp training data,
especially for multi-fingered grasping. Representative approaches to
grasp learning for  parallel jaw grippers collect tens of
thousands~\cite{pinto2016supersizing}, hundred
thousands~\cite{levine2016learning}, millions~\cite{mahler2017dex}, or even
one billion~\cite{eppner2019billion} grasp attempts.

Compared with grasp training data of two-fingered parallel jaw
grippers, it is more difficult to collect grasping data for
multi-fingered hand. Only $3,770$ grasps were generated for the
Barrett hand using GraspIt!~\cite{ciocarlie2007dexterous}
in~\cite{varley2015generating}, Lu et al.~\cite{lu2017grasp}
collected only $1507$ grasp attempts for the four-fingered Allegro
hand, and Veres et al.~\cite{veres2017modeling} collected a data-set
of around $47,000$. In this paper, we propose a novel active grasp
learning approach in order to address the data collection bottleneck
for grasp learning on multi-fingered hands in order to avoid simply
scaling up grasp attempts to significantly larger numbers.

Reinforcement learning stands as an alternative to supervised learning
techniques for robotic grasping~\cite{kalashnikov2018qt,
  wu2020generative}. However, the sample complexity of reinforcement learning generally
requires even more data than supervised approaches as the robot must
learn not only the necessary grasp configuration, but also how to
reach and lift the target object.
For example Kalashnikov et al.~\cite{kalashnikov2018qt} had to collect
over 580k grasps over the course of several weeks across 7 robots to
learn to grasp with a two-fingered gripper. An alternative is to
leverage human demonstrations to provide an initial supervised
learning phase to the RL process, at the cost of giving up full
autonomous exploration and labeling~\cite{osa2016experiments}.

Active learning scenarios can be organized~\cite{settles2012active} as: 1) stream-based
selective sampling where a stream of samples come in and the learner
selects if the current observed sample should get a label; 2)
pool-based sampling where the learner has a fixed set of unlabeled
samples and can ask which sample can next get a label; and 3) query
synthesis where the learner selects what input sample to generate and
query for the label.

Kroemer et al.~\cite{kroemer2010combining} combine active learning and reactive control for robot grasping. 
They propose a hierarchical controller to decide where and how to grasp the object. The upper level controller selects where to grasp the object using a reinforcement learner, which models the reward distribution over the action space using Gaussian Process Regression (GPR). Active learning is applied to incorporate the supervised data into GPR. The lower level controller comprises an imitation learner and a vision-based reactive controller to determine appropriate grasping motions. 
Compared with our work, their grasp learning and planning are performed separately for each given object without sharing information across different objects. Moreover, they only model the 6 DOF hand pose without considering the hand joint angles. 
In~\cite{morales2004active}, an active learning approach is presented
for assessing robot grasp reliability. 
It classifies grasp success using the weighted \(k\)-nearest neighbors
algorithm and defines a classification confidence metric for their pooling-based active learning. 
However, the grasp active learning in~\cite{morales2004active} is evaluated as an offline classification problem without performing grasp planning.  

Montesano et al.~\cite{montesano2012active} address the problem of
actively learning good grasping points using a pooling formulation
to reduce the number of examples needed for training the grasping
model. This paper uses a non-parametric kernel approach to compute the
success probability of grasping points. Its proposed active learning
metric favors exploration of areas with high probability of success or where uncertainty is high.
This pooling-based active learning is tested on a real humanoid robot for both seen and unseen objects. 
In~\cite{fu2019active}, active perception is used to collect the training data set efficiently for each specific object, although no active learning is performed
over grasp configuration or control.  
Tian et al.~\cite{tiantransferring} compute an approximation of the grasp space for an object of interest using SVM-based active learning and apply bijective contact mapping to transfer grasp contact points from an example object to a novel object. However, their active learning focuses on the separation of in-collision grasp configurations from the collision free configurations, instead of the grasp quality.

Existing bandit-based grasping work~\cite{kroemer2010combining, laskey2015multi, mahler2016dex} treats each possible grasp as one arm and perform grasp planning separately for different objects. Instead, we use the bandit-framework to select qualitatively different grasps to actively improve our grasp deep learning models in a unified manner across all objects in our training set. Moreover, none of these multi-armed bandit grasp planning work focus on rigid multi-fingered hand grasping like we do.
In~\cite{kroemer2010combining}, active grasp learning for the Barrett
hand is formulated as a continuum-armed bandits problem by treating
each grasp pose as one arm. A continuum Gaussian Bandits algorithm is
proposed that identifies the local maxima using a gradient-based method inspired by mean shift.

In~\cite{laskey2015multi, laskey2015budgeted}, multi-armed bandit is used to generate force closure grasps for parallel jaw grippers on each given 2D planar object. 
Mahler et al.~\cite{mahler2016dex} extends the multi-armed bandit grasp planning of~\cite{laskey2015multi} from 2D to 3D objects for parallel jaw grippers. The similarity between a pair of grasps and objects is measured and used as the prior information for the multi-armed bandit grasp planning. The object feature is extracted using a Multi-View CNN. Thompson sampling is used to solve the grasping multi-armed bandit problem in~\cite{laskey2015multi, laskey2015budgeted, mahler2016dex}.
Oberlin et al.~\cite{oberlin2018autonomously} formalize grasping for parallel jaw grippers as a multi-armed bandit problem. It defines a new algorithm called Prior Confidence Bound for best arm identification in budgeted bandits, which enables the robot to quickly find an arm corresponding to a good grasp without pulling all the arms. A linear grasp model is used to score grasp candidates for their grasping multi-armed bandit. They take an instance-based approach that needs to collect training data for all testing objects to be grasped, which does not generalize to novel objects.

In~\cite{eppner2017visual}, a contextual multi-armed bandit formulation selects one of three pre-defined environment-constrained grasping strategies for a given object, instead of learning the grasp configuration across different objects as we do. Different multi-armed bandit algorithms such as Thompson sampling and GP-UCB are compared. 

Compared with existing grasp active learning work, our active learning approach has three novelties: the first grasp active learning work leveraging deep networks; doing a continuous synthesis approach instead of pooling for grasp active learning; modeling grasp active learning as a multi-armed bandit problem to cover the grasp space across different objects efficiently.



\section{Grasp Planning as Inference}
\label{sec:grasp_inf}
Our grasping method has two stages. In the learning stage described in
Section~\ref{sec:active_learning}, we use active learning to train the
grasp model comprised of the voxel-based classifier and the mixture
density network (MDN) prior. In the planning stage, we perform maximum
a posteriori (MAP) inference using this model to synthesize a grasp
preshape configuration for execution on the robot.

In this work, we define the
grasp configuration as the palm pose and the hand's preshape joint angles that define the
shape of the hand prior to closing~\cite{lu2020multi}.
In order to make the grasp inference agnostic to object poses, we
encode the palm pose in the object reference frame for learning. At
inference time we optimize over the robot arm joint configuration
solving the arm inverse kinematics jointly with the grasp configuration.
After finding the grasp preshape configuration, the robot moves to
this preshape and runs a controller to close the hand forming the
grasp on the object. The specific joints that define the preshape can
be found in~\cite{lu2020multi}.

We focus on scenarios where a single, isolated object of interest is present in the scene. Importantly, we assume no explicit knowledge of the object beyond a single camera sensor reading of it in its current pose.
The problem we address is, given such a grasp scenario, plan a grasp preshape configuration that allows the robot to successfully grasp and lift the object without dropping it.

Given the learned model parameters, \(\bm W\) and \(\bm\Phi\), along
with the visual representation, \(\bm z\), associated with an observed
object of interest, our goal is to infer the grasp configuration \(\bm q\) in the robot arm joint configuration space that maximizes the posterior probability of grasp success \(Y=1\). Here \(Y\) defines a random Boolean variable with 0 meaning failure and 1 meaning success.
We can thus formalize grasp planning as a \emph{maximum a posteriori} (MAP) inference problem:
\begin{align}
\begin{split}
& \underset{\boldsymbol{q}}{\text{argmax}} \quad
p(\boldsymbol{q} | Y=1, {\bm z}, \bm{W}, \bm\Phi) \propto p(Y=1 | \boldsymbol{q}, z, \bm{W}) p(\boldsymbol{q} | z, \bm\Phi) \\
& \text{subject to}\hspace{8pt} {\bm q}_{min} \leq \bm q \leq {\bm q}_{max}
\end{split}
\label{eq:grasp_inf}
\end{align}
We constrain the grasp configuration parameters to obey the joint
limits of the robot hand in Eq.~(\ref{eq:grasp_inf}). Other constraints
such as collision avoidance could optionally be added~\cite{vandermerwe-icra2020-reconstruction-grasping}.

\begin{figure}
	\centering
	\begin{subfigure}[]{0.45\textwidth}
		\includegraphics[width=\textwidth]{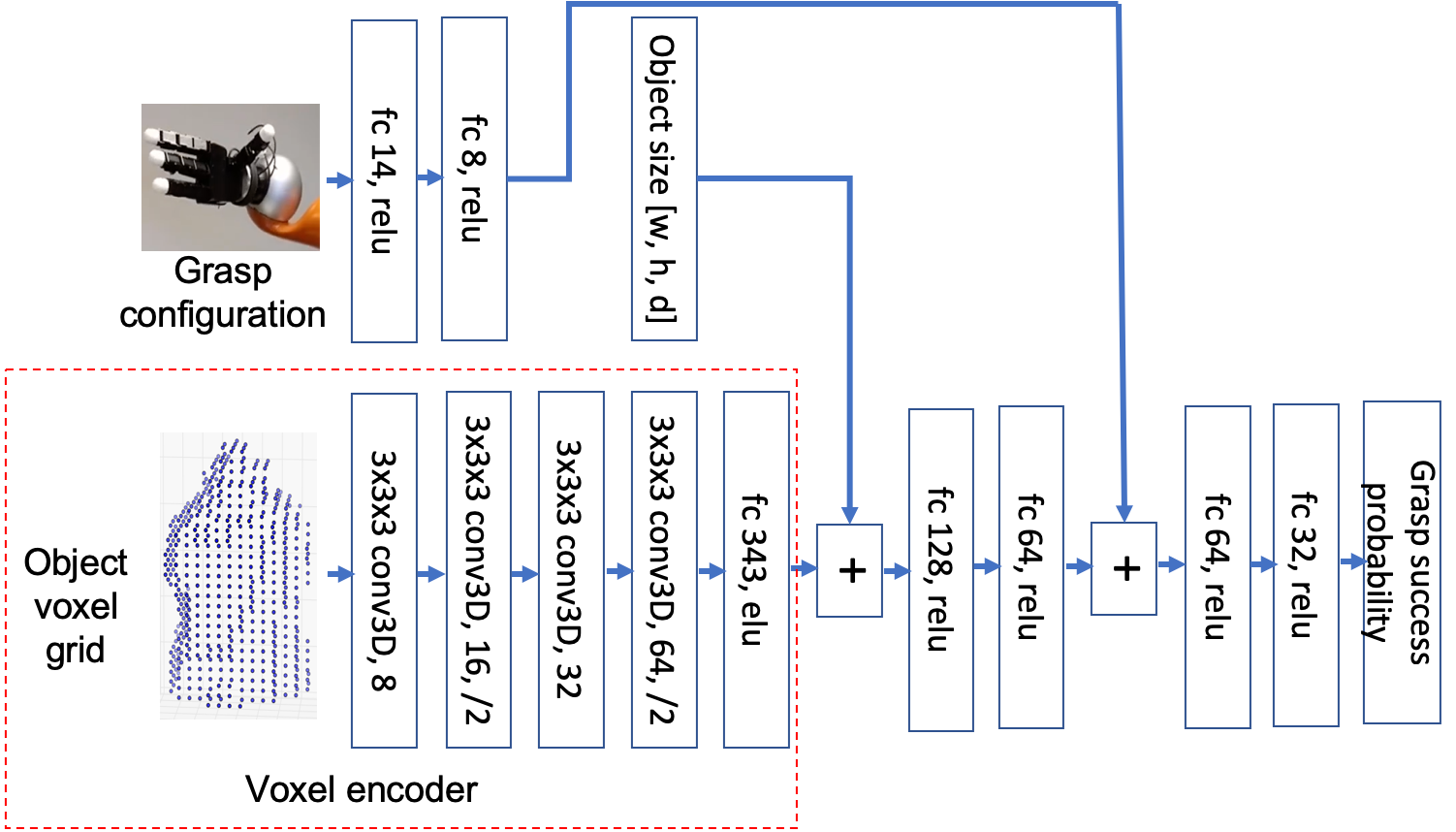}
		\caption{The architecture of our grasp success classifier.}
		\label{fig:voxel-config-net}
	\end{subfigure}
	\begin{subfigure}[]{0.45\textwidth}
		\includegraphics[width=\textwidth]{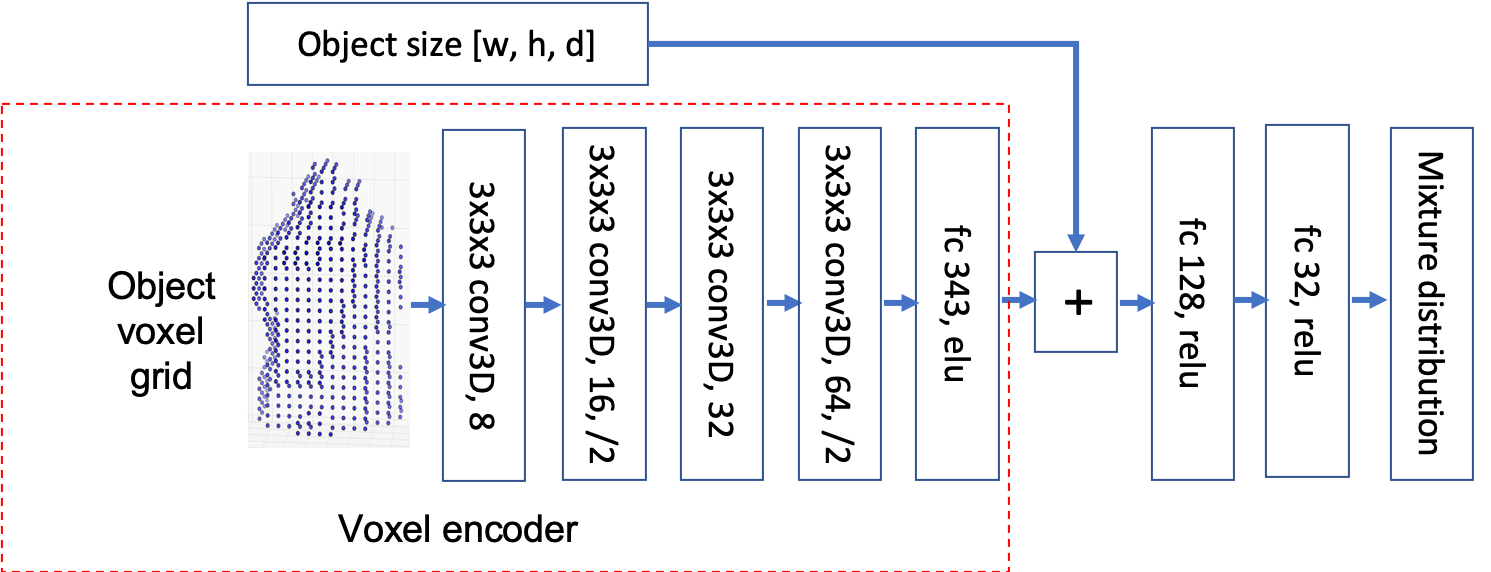}
		\caption{The architecture of our grasp conditional prior.}
		\label{fig:mdn_arch}
	\end{subfigure}
	\caption{The voxel-config-net and MDN architectures~\cite{lu2020multi}. Bottom left visualizes the voxel-grid for the ``mustard bottle'' object. All convolutional layers use $3 \times 3 \times 3$ 3D convolutional filters with exponential linear unit (ELU) activations. We annotate the number of filters and the stride ($/2$ means a stride of 2) for convolutional layers. We annotate the number of neurons and activation function for fully connected layers.}
	\label{fig:voxel-nets}
	\vspace{-10pt}
\end{figure}

We define the grasp success likelihood \(p(Y=1 | {\bm q}, \bm{z}, \bm{W})\) to be a voxel-based 3D convolutional network following~\cite{lu2020multi}. 
\(\bm{W}\) represents the neural network parameters. The voxel-based
network predicts the probability of grasp success, \(Y\), as a
function of the visual representation of the object of interest,
\(\bm z\), and grasp configuration, \(\bm q\). Figure~\ref{fig:voxel-config-net} shows the architecture of our grasp success prediction network.

We additionally learn a model of the conditional probability
distribution \(p(\bm{q} | \bm{z}, \bm{\Phi})\) encoding
the distribution of successful grasps given the object of interest. We
construct this as a mixture-density network (MDN) following~\cite{lu2020multi}. Given its input an MDN predicts the
parameters (means, covariance and mixing weights) of a Gaussian
mixture model as output. \(\bm\Phi\) define the learned
weights of the MDN. Figure~\ref{fig:mdn_arch} shows the architecture of our grasp prior. We train our voxel-based MDN using the negative log likelihood loss.

We use the object voxel-grid and the object size vector as the visual object representation~\cite{lu2020multi}. In order to generate the voxel-grid we first segment the object from the 3D point cloud by fitting a plane to the table using RANSAC and extracting the points above the table. We then estimate the first and second principle axes of the segmented object to create a right-handed object reference frame aligned relative to the world frame. We compute the object size along the three coordinates of the object reference frame to construct the object size vector. 
We then generate a $32 \times 32 \times 32$ voxel grid oriented about this reference frame. More details of the voxel-grid generation are described in~\cite{lu2020multi}.

We solve the grasp inference in the log-probability space and
regularize the log-prior with a multiplicative gain of $0.5$ to
prevent the prior dominating the inference. We use the scikit-learn
implementation of the popular L-BFGS optimization algorithm with bound constraints to efficiently solve the
inference problem.\footnote{http://scikit-learn.org/stable/index.html} We initialize the optimization by sampling from the MDN prior.


\section{Active Grasp Learning}
\label{sec:active_learning}
As explained in Section~\ref{sec:related_work} three primary active
learning scenarios have been considered in the literature: stream-based selective sampling, pool-based sampling, and query
synthesis~\cite{settles2012active}. In this paper, we examine the
query synthesis approach in a continuous fashion for active grasp learning.


Our active learning approach combines three separate exploration
strategies: the likelihood uncertainty of the grasp success
classifier, grasp success probability maximization, and the grasp
configuration exploration. These three arms are designed to improve
both the grasp success classifier and enable the grasp model to cover
the space of grasp configuration as much as possible.

We formulate active learning as a multi-arm bandit problem to balance exploration and exploitation
across our three different strategies. The multi-armed bandit is a classic
reinforcement learning problem where we are given $n$ slot machines
(arms)~\cite{Sutton2018a}.
Each slot machine has its own probability distribution of reward. At each
time step, the agent chooses one slot machine to play and receives a
reward. The agent's objective is to decide which arm to play at each time step such that it can maximize the cumulative reward.

\newcommand\mycommfont[1]{\footnotesize\ttfamily\textcolor{blue}{#1}}
\SetCommentSty{mycommfont}
\begin{algorithm}
	\SetAlgoLined
 	
 	\KwData{The geometrical grasp data $GeoData$ collected using a heuristic geometrical planner.}
 	\KwResult{The grasp model $G$, including the voxel-based classifier and the MDN conditional prior.}
	Initialize $G$ through supervised learning using $GeoData$\;
	$N=128, M=16, K=4, E=5$\;
	$i = 1$\;
	Initialize the active learning grasping data $ActiveData$ to be an empty dataset\; 
	\tcp{$N$: Number of active learning rounds}
	\While{$ i \leq N$}{
		$j = 0$\;
		Initialize $CurRoundData$ to be an empty dataset\;
		\tcp{$M$: Number of batches for an active learning round}
		\While{$j < M$}{ 
			Synthesize an active learning grasp query $Q$ using the multi-arm bandit algorithm UCB\;
			Add $Q$ into $CurRoundData$\;
			$j = j + 1$\;
		}
		Add $CurRoundData$ into $ActiveData$\;
		\tcp{$K$: For every $K$ rounds, online update using both $GeoData$ and $ActiveData$}
		\eIf{$i \% K == 0$}{
			Online update the grasp model $G$ using both $GeoData$ and $ActiveData$ for $E$ epochs\;
		}{
			Online update the grasp model $G$ using $CurRoundData$ for $E$ epochs\;
		}
		$i = i + 1$\;
	}
	\caption{Pseudocode of our multi-armed bandit active learning.}
	\label{alg:active_pseudocode}
\end{algorithm}


\subsection{Active Grasp Learning Strategies}
\textbf{The grasp likelihood uncertainty maximization arm} defines a
standard active learning objective for binary classification problems
in order to improve classification accuracy.
We define the likelihood uncertainty maximization arm in Eq.~(\ref{eq:max_clf_uncertainty}). 
\begin{align}
\begin{split}
& \underset{\boldsymbol{q}}{\text{argmax}} \quad f(\boldsymbol{q}) + g(\boldsymbol{q}) \\
& \text{subject to}\hspace{8pt} {\bm q}_{min} \leq \bm q \leq {\bm q}_{max} \\
\end{split}
\label{eq:max_clf_uncertainty}
\end{align}

\begin{align}
\begin{split}
& f(\boldsymbol{q}) = \frac{1}{2 ( 1 + \exp(-\log p(\boldsymbol{q} | z, \bm\phi)))} \\
\end{split}
\label{eq:uct_f}
\end{align}

\begin{align}
\begin{split}
& g(\boldsymbol{q}) = \left\{ \begin{array}{cc} 
p(Y=1 | \boldsymbol{q}, z, \boldsymbol{w}) & p(Y=1 | \boldsymbol{q}, z, \boldsymbol{w}) <= 0.5 \\
1 - p(Y=1 | \boldsymbol{q}, z, \boldsymbol{w}) & p(Y=1 | \boldsymbol{q}, z, \boldsymbol{w}) > 0.5
\end{array} \right. \\
\end{split}
\label{eq:uct_g}
\end{align}

Given the learned model parameters \(\bm W\), \(\bm\Phi\), and the
visual representation \(\bm z\) associated with an observed object of
interest, our goal is to infer the grasp configuration parameters $\bm
q$ that maximize the uncertainty of the grasp success likelihood
\(Y\). 
$f(\boldsymbol{q})$ in Eq.~(\ref{eq:uct_f}) represents the grasp
success classification uncertainty. $g(\boldsymbol{q})$, defined in Eq.~(\ref{eq:uct_g}), regularizes the optimization to not stray into areas far from grasp configurations observed in the training data. 
We use L-BFGS with bound constraints to solve the uncertainty maximization initialized with a configuration sampled from the prior. We treat its optimized objective function value as its bandit reward. 

\textbf{The grasp success probability maximization arm} encourages the
active learner to synthesize more successful grasps to overcome the
common issue of class imbalance in multi-fingered grasp learning.
For example in~\cite{lu2017grasp} only $11\%$ out of $1507$ training
grasps generated using a geometric grasp planner were successful. 
We perform the same grasp inference as in Eq.~(\ref{eq:grasp_inf}) for
this arm. We normalize the logarithm of the optimized grasp success
posterior with a Sigmoid function as the reward for the bandit. 

\textbf{The grasp configuration exploration arm} enables the grasp
model to cover a larger portion of the grasp configuration space by
exploring the areas farther from previous grasp attempts. To achieve
this, we sample $50$ grasp configuration candidates from the prior and select the one with lowest prior probability density as the grasp to explore. We compute the Sigmoid of the negative logarithm of the prior density as the reward of the selected grasp.

\subsection{Implementation Details}
During training we randomly select objects to present to the active learner in order to cover the
object space well. We solve the bandit problem using the classical Upper Confidence Bound (UCB) algorithm~\cite{Sutton2018a}. The UCB algorithm exploits actions with high average rewards obtained and explores more uncertain actions. 
Algorithm~\ref{alg:active_pseudocode} shows the pseudocode of our
multi-armed bandit active learning. 
Online training is a known difficult problem for active deep learning~\cite{gal2017deep}.
For each round of active learning, we first apply the multi-arm bandit algorithm to generate an active learning mini-batch that contains $16$ grasps using the model of previous round (i.e. data acquisition). Then we online update the grasp model for each round, including the voxel-based classifier and the MDN prior. Depending on the active learning round number, we use either the mini-batch data of the current  round or all the active and geometrical data for online training. 



\section{Experiments}
\label{sec:experiments}
In this section, we describe the experimental evaluation and analysis
of our grasp active learner. We compare our active learning approach
to passive supervised learning approach used to initialize our active
learning grasp model. We then compare to the passive supervised
learning model trained with more samples.

\subsection{Robotic System for Data Collection and Experiments}
\label{subsec:robot_sys}
\begin{figure}[h]
	\centering
	\includegraphics [width=0.45\textwidth] {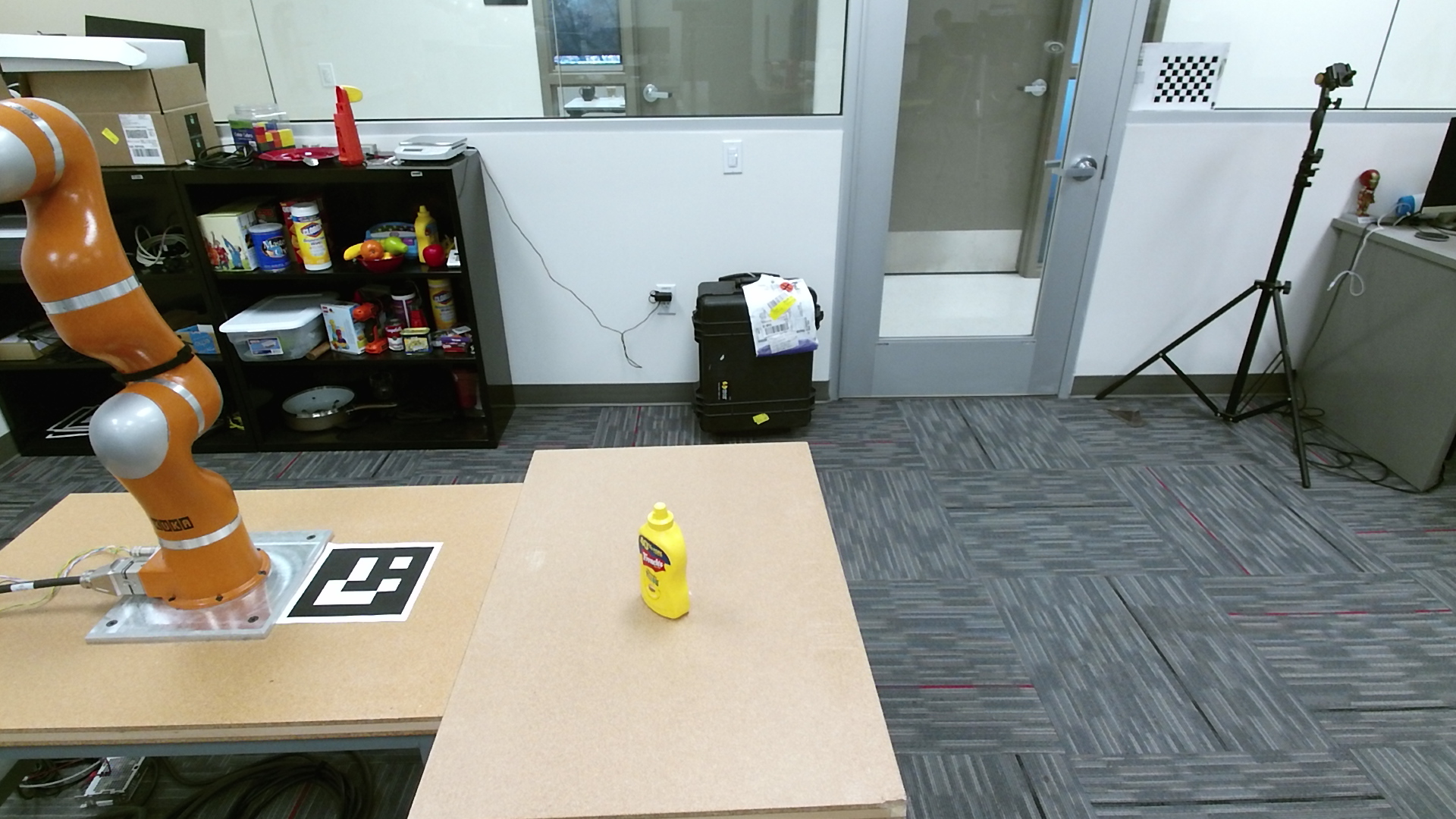}
	\caption{Example RGB image of the experimental robot setup from the RGB-D Kinect 2 camera.}
	\label{fig:kinect2_image}
	\vspace{-12pt}
\end{figure}

We conduct all training and experiments using the four-fingered, 16 DOF Allegro hand mounted on a Kuka LBR4 7 DOF arm. We evaluate our grasp planners on the physical robot.
There are 15 parameters for the Allegro hand preshape, 7 for the LBR4 arm joint angles representing the palm pose and 8 relating to the first 2 joint angles of each finger proximal to the palm. 
We use a Kinect2 camera to generate the point cloud of the object on the table. 
One example RGB image of the robot and the object generated by Kinect2 can be seen from Figure~\ref{fig:kinect2_image}. 


\subsection{Active Learning Setup and Analysis}
\label{subsec:active_analysis}
We train the initial grasp classifier and MDN prior, for active
learning in a passive supervised way using the data
from~\cite{lu2020multi}. These data contain $10$ sets of grasps on the
Bigbird objects data set~\cite{singh2014bigbird} containing a total of
$8,988$ grasps, $2308$ of which were successful. We randomly select 5 sets (i.e. $4,578$ grasp samples) out of the 10 training sets in~\cite{lu2020multi} to train the initial grasp model for active learning. $1,168$ out of these $4,578$ grasps are successful grasps.

We run active learning for $128$ rounds in simulation. Our active learner generates $16$ grasp queries per round.
We online update the grasp model for each round of active
learning as described in Section~\ref{sec:active_learning}. We
randomly select a different Bigbird object every $5$ active learning queries in order to cover the object space well. 

\begin{table}[]
	\begin{tabular}{|l|l|l|l|}
		\hline
		& Max success & Max uncertainty & Exploration \\ \hline
		Average reward  & 0.965       & 0.922           & 0.933       \\ \hline
		Average time  & 8.11       & 4.15           & 2.53       \\ \hline
		Number of pulls & 933         & 522             & 597        \\ \hline
	\end{tabular}
	\caption{Average rewards and number of pulls for the $3$ active learning arms.}
	\label{tab:active_stats}
	\vspace{-15pt}
\end{table}

The average rewards, average running time, and number of plays of our three active learning arms can be seen from Table~\ref{tab:active_stats}.
We empirically add a constant of $0.35$, $-0.05$, and $0.6$ to the max
success, max uncertainty, and exploration arm rewards, respectively, for the multi-armed bandit active learning. 
The constant terms cause the rewards of different arms to be in a similar range.

\begin{figure}[h]
	\centering
	\includegraphics[ width=0.9\linewidth]{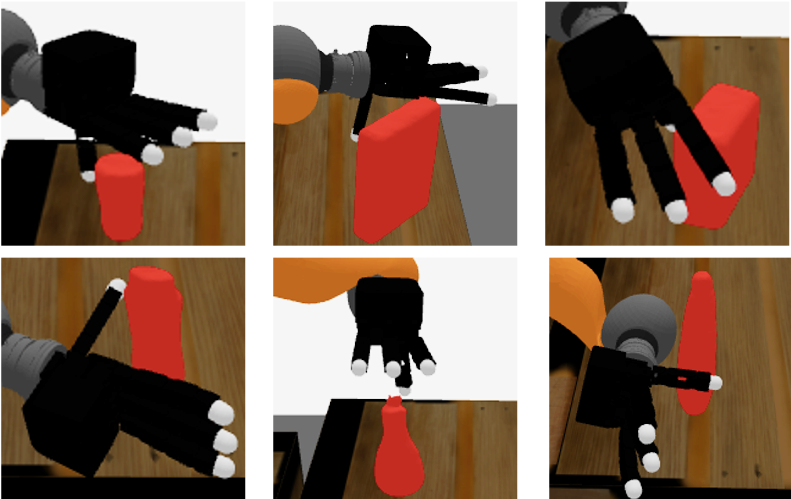}
	\caption{Examples of grasp preshapes queries generated by our three active learning arms. The columns from left to right show grasp queries synthesized by the max success, max uncertainty, and the exploration arm respectively.}
	\label{fig:grasp_arm_examples}
	\vspace{-8pt}
\end{figure}
\begin{figure}[h]
	\vspace{-5pt}
	\centering
	\includegraphics[ width=1.\linewidth]{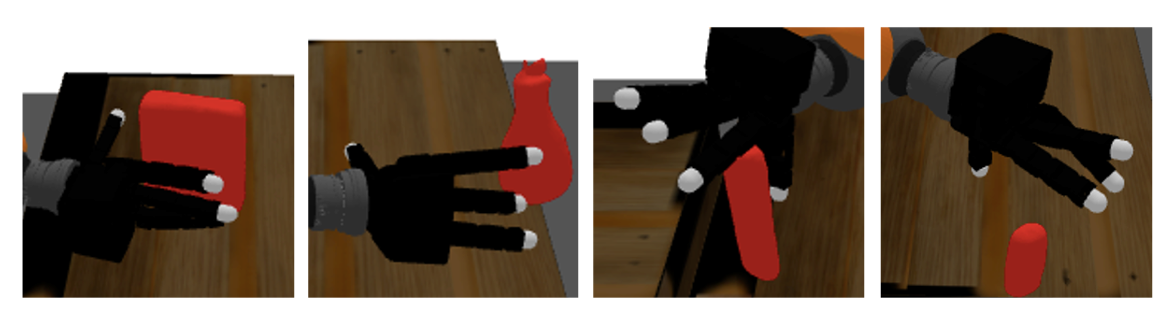}
	\caption{Active grasp preshape examples that the heuristic grasp planner during data collection is not able to generate.}
	\label{fig:active_grasp_low_heu_examples}
	\vspace{-5pt}
\end{figure}

Figure~\ref{fig:grasp_arm_examples} shows example grasps queried by
the three different arms of our active learner.
By measuring the differential entropy~\cite{jost2006entropy, spellerberg2003tribute} of the collected training data,
we see that the active learner generates a larger diversity  of grasp preshape configurations than the heuristic
planner. 
The grasp palm pose for entropy computation is in Cartesian space, while hand joint angles are in radians.
We generate $4$ heuristic grasp subsets for entropy computation by randomly selection without replacement. 
Each subset has the same number of grasps as the active dataset.
Table~\ref{tab:active_learn_entropy} summarizes these results.
\begin{table}[]
	\begin{tabular}{|l|l|l|l|}
		\hline
		& Active learning & Heuristic mean & Heuristic std \\ \hline
		Config entropy  & $-6.1$       & $-33.6$
		& $0.05$ \\ \hline
		Pose entropy  & $0.1$       & $-1.5$
		& $0.05$ \\ \hline
		Joint entropy  & $-6$       & $-32.1$
		& $0.007$ \\ \hline
	\end{tabular}
	\caption{The differential entropy of multivariate Gaussian distributions fit to active learning and heuristic grasping data.}
	\label{tab:active_learn_entropy}
		\vspace{-10pt}
      \end{table}
      In Figure~\ref{fig:active_grasp_low_heu_examples}, we show
      preshape examples from the active learner which the heuristic
      grasp planner would not able to generate.
      
\subsection{Real Robot Experiments}
We compare our active learning model with the initial passive
supervised model to examine if our active learner can improve on the
initial model. We also compare our active learning model with a
passively learned model trained on $8,988$ grasps
from the dataset of~\cite{lu2020multi} using the same learning
criteria as in that paper.

\begin{figure}[t!]
	\centering
	\includegraphics[ width=0.9 \linewidth]{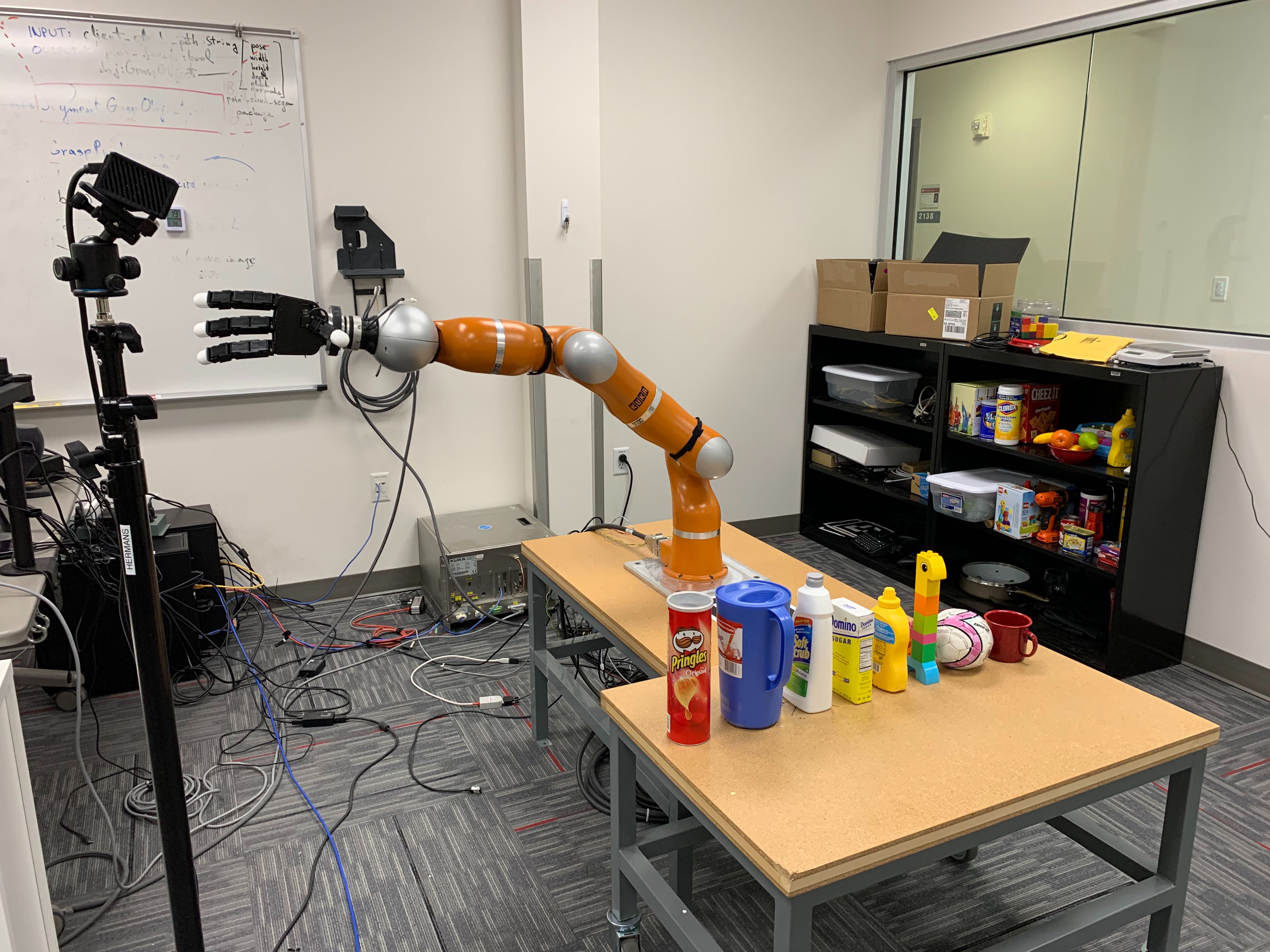}
	\caption{Experimental setup with objects used for experiments. From left to right objects are ``pringles'', ``pitcher'', ``soft scrub'', ``sugar box'', ``mustard bottle'', ``Lego'', ``soccer ball'', and ``mug''. Objects range in size from \(8 \times 9 \times 11\)cm (mug) to \(13 \times 17 \times 24\)cm (pitcher).}
	\label{fig:exp_setup}
	\vspace{-12pt}
\end{figure}

We evaluate grasp planning as inference using the active learning model, the passive supervised model trained with $8,988$ geometrical grasps, and the passive supervised initial model trained with $4,578$ geometrical grasps on the physical robot system.
We perform experiments on 8 YCB~\cite{calli2015benchmarking} objects covering different textures, shapes, and sizes. We show the experimental setup and objects used in Figure~\ref{fig:exp_setup}.  All experimental objects are unseen in training except for ``Pringles''. We attempted grasps at $5$ different poses per object, for a total of $40$ grasp attempts per method. We use the same set of locations across different methods, but each object has its own set of random orientations. In total, we performed $120$ grasp attempts for $3$ different methods across $8$ objects in this paper. 

If the RRT-connect motion planner fails to generate a plan for a grasp
due to collision avoidance, we generate a new grasp using the same
grasp planner with a different initialization. If the grasp planner
could not generate a grasp with a motion plan in $5$ attempts, we
treat the grasp attempt as a failure case. It turns out all three
grasp planners can find a grasp with a feasible motion plan for every
object pose we tested in $5$ attempts for all objects but the mug where
no method could find a successful side grasp.


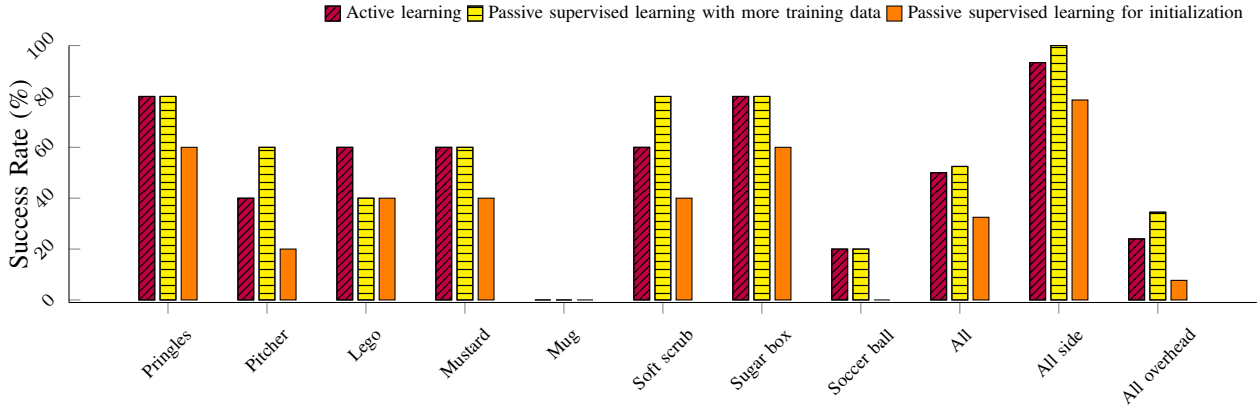
\begin{figure*}[ht!]
	\begin{adjustbox}{minipage={\columnwidth}}
		\begin{tikzpicture}
		\begin{axis}[
		ybar, ymin=-1, ymax=100,
		ylabel={Success Rate (\%)},
		axis lines*=left,
		symbolic x coords={Pringles, Pitcher, Lego, Mustard, Mug, Soft scrub,  Sugar box, Soccer ball, All, All side, All overhead},
		xtick=data,
		ticklabel style = {font=\scriptsize, rotate=45},
		legend style={font=\scriptsize, draw=none, fill=none},
		y label style={at={(axis description cs:0.02,0.5)},anchor=north},
		bar width = 6pt, height=5cm, width=2*\linewidth,
		legend style={area legend, at={(1,1.20)}, anchor=north east, legend columns=3, },
		legend image code/.code={%
			\draw[#1] (0cm,-0.1cm) rectangle (2mm,1mm);},
		]
		\addplot [fill=purple, postaction={pattern=north east lines}] coordinates {
			(Pringles, 80) (Pitcher, 40) (Lego, 60) (Mustard, 60) (Mug, 0) (Soft scrub, 60) (Sugar box, 80) (Soccer ball, 20)
			(All, 50) (All side, 93.3) (All overhead, 24)};
		
		\addplot [fill=yellow, postaction={pattern=horizontal lines}] coordinates {
			(Pringles, 80) (Pitcher, 60) (Lego, 40) (Mustard, 60) (Mug, 0) (Soft scrub, 80) (Sugar box, 80) (Soccer ball, 20)
			(All, 52.5) (All side, 100) (All overhead, 34.5)};
		
		\addplot [fill=orange] coordinates {
			(Pringles, 60) (Pitcher, 20) (Lego, 40) (Mustard, 40) (Mug, 0) (Soft scrub, 40) (Sugar box, 60) (Soccer ball, 0)
			(All, 32.5) (All side, 78.6) (All overhead, 7.7)};
		\legend{Active learning, Passive supervised learning with more training data, Passive supervised learning for initialization}
		\end{axis}
		\end{tikzpicture}
	\end{adjustbox}
	\caption{Multi-fingered grasping success rates of grasp inference using 3 different models on the real robot. ``Pringles'' was seen in training, other $7$ objects are previously unseen.}
	\label{fig:active_grasp_results}
\end{figure*}

\begin{figure*}[h]
	\vspace{-8pt}
	\centering
	\includegraphics[ width=0.9\linewidth]{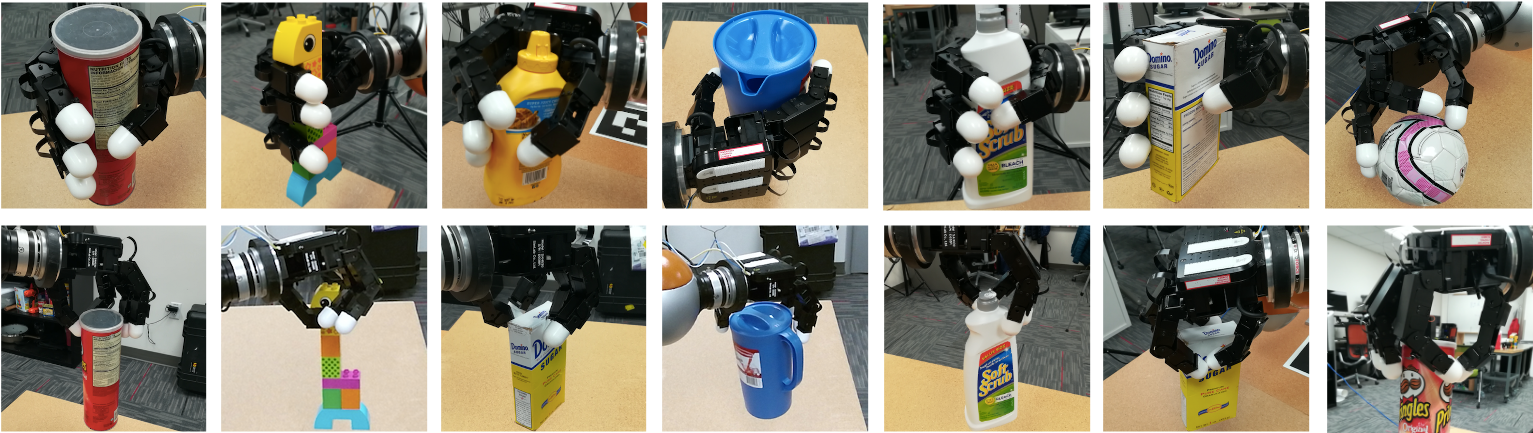}
	\caption{Examples of successful grasps generated by inference with our actively learned grasp model. The top row shows side grasps. The bottom row shows overhead grasps.}
	\label{fig:active_grasp_examples}
	\vspace{-10pt}
\end{figure*}

As described in Section~\ref{subsec:robot_sys}, we label a grasp attempt that successfully lifts the object to a height of $0.15$m without dropping it as successful. We also manually label each experimented grasp to be a side or overhead grasp. 

The grasp success rates for all three methods are summarized in Figure~\ref{fig:active_grasp_results}. 
It takes around $3-10$ seconds for each method to generate a grasp.
The grasp planners using the active learning model, the passive supervised model with more training data, and the passive supervised model for initialization achieve grasping success rates of $50\%$, $52.5\%$, and $32.5\%$ respectively for the $8$ objects.

The grasp inference using the active learning model generated $15$
side and $25$ overhead grasps for the $8$ test objects. The grasp
inference using the passive supervised model with more training data
plans $11$ side and $29$ overhead grasps, while the smaller data
passive model generated $14$ side and $26$ overhead grasps.
More overhead grasps than side grasps were generated for all three
methods both in training and experiments. Overhead grasps are relatively further away from the table compared with side grasps,
which makes it easier for the motion planner to avoid collision with the table. 

We also report the grasp success rates of side and overhead grasps separately for each grasp planner in Figure~\ref{fig:active_grasp_results}. 
The grasp inference using the active learning model, the passive supervised model with more training data, and the passive supervised model for initialization achieve success rates of $93.3\%$, $100\%$, and $78.6\%$ respectively for side grasps of the $8$ objects.
The grasp inference using the active learning model, the passive supervised model with more training data, and the passive supervised model for initialization achieve success rates of $24\%$, $34.5\%$, and $7.7\%$ respectively for overhead grasps of the $8$ objects.

All three grasp planners have lower success rate for overhead grasps than side grasps. Objects such as mustard, mug, and lego have relatively smaller contact areas available for overhead grasps and our grasp controller would push them away when closing the hand as it had no feedback from vision or haptic sensors to know the object was moving~\cite{lu2020multi}. 

The grasp planner using the actively learned model outperforms the passive supervised model for initialization, which shows that active learning improves the supervised passive model for grasp inference. 
The grasp planner using the active learning model achieves comparable
performance with the passive supervised learning with fewer grasp
samples, which demonstrates the benefit of grasp active learning. This
implies that our active grasp learning covers the grasp configuration
space better across different objects with fewer samples, compared
with passive supervised grasp learning. As noted above no planner
could find a side grasp for the mug; additionally, all overhead grasp
attempts for the mug failed during execution.

In Figure~\ref{fig:active_grasp_examples}, example grasps are shown for different objects generated by our inference approach with the actively learned grasp model, including the voxel-based classifier and MDN prior. We show side grasps that provide stability in the top row. We present overhead grasps in the bottom row, which offer dexterity and access to objects in clutter.
\begin{table}[]
	\begin{tabular}{|l|l|l|l|}
		\hline
		& Active & Supervised more & Supervised less \\ \hline
		Config entropy  & $-16.3$       & $-18.6$
		& $-15.9$ \\ \hline
		Pose entropy  & $-3.7$       & $-4.2$
		& $-3.7$ \\ \hline
		Joint entropy  & $-11.8$       & $-13.5$
		& $-11$ \\ \hline
	\end{tabular}
	\caption{The differential entropy of a multivariate Gaussian
          distribution fit to our real-robot experiment grasps for the three different models.}
	\label{tab:active_exp_entropy}
	\vspace{-15pt}
\end{table}
Lastly, Table~\ref{tab:active_exp_entropy} shows the active learner plans real-robot grasps with greater
diversity than the passively learned large-data model.


\section{Discussion and Conclusion}
\label{sec:discussion}
In this work, we propose an approach to active multi-fingered, grasp
learning using three different exploration strategies. We formalize
these strategies as arms in a bandit setting. Our real-robot grasping
experiment shows our active grasp planner trained using fewer data
achieves comparable success rates with a passive supervised planner
trained on data generated by a geometrical grasp planner. This implies our active grasp learning covers the grasp configuration space better across different objects with fewer samples, compared with passive supervised grasp learning. We also compute the differential entropy to demonstrate our active learner generates grasps with larger diversity than passive supervised learning using more heuristic data, which attains comparable success rate.

In the future, we will run more rounds of active learning in order to
cover more of the grasping and objects space, which can help to
generate grasps with more diversities for different tasks. We plan to
extend our approach to also actively select which object from a set to
attempt a grasp on. We believe learning or designing a more complex
feedback controller for overhead grasps using tactile feedback would
boost the overhead grasp performance~\cite{lu2020multi}. 

%



\bibliographystyle{IEEEtran}
\bibliography{references}  

\begin{thebibliography}{10}
\providecommand{\url}[1]{#1}
\csname url@samestyle\endcsname
\providecommand{\newblock}{\relax}
\providecommand{\bibinfo}[2]{#2}
\providecommand{\BIBentrySTDinterwordspacing}{\spaceskip=0pt\relax}
\providecommand{\BIBentryALTinterwordstretchfactor}{4}
\providecommand{\BIBentryALTinterwordspacing}{\spaceskip=\fontdimen2\font plus
\BIBentryALTinterwordstretchfactor\fontdimen3\font minus
  \fontdimen4\font\relax}
\providecommand{\BIBforeignlanguage}[2]{{%
\expandafter\ifx\csname l@#1\endcsname\relax
\typeout{** WARNING: IEEEtran.bst: No hyphenation pattern has been}%
\typeout{** loaded for the language `#1'. Using the pattern for}%
\typeout{** the default language instead.}%
\else
\language=\csname l@#1\endcsname
\fi
#2}}
\providecommand{\BIBdecl}{\relax}
\BIBdecl

\bibitem{bohg2014data}
J.~Bohg, A.~Morales, T.~Asfour, and D.~Kragic, ``{Data-Driven Grasp Synthesis-a
  Survey},'' \emph{{IEEE} Trans. on Robotics}, vol.~30, no.~2, pp. 289--309,
  2014.

\bibitem{saxena2006learning}
A.~Saxena, J.~Driemeyer, and A.~Y. Ng, ``{Robotic Grasping of Novel Objects
  using Vision},'' \emph{Intl. Journal of Robotics Research}, vol.~27, no.~2,
  pp. 157--173, 2008.

\bibitem{Saxena-aaai2008}
A.~Saxena, L.~L.~S. Wong, and A.~Y. Ng, ``{Learning Grasp Strategies with
  Partial Shape Information},'' in \emph{AAAI National Conf. on Artificial
  Intelligence}, 2008, pp. 1491--1494.

\bibitem{lenz2015deep}
I.~Lenz, H.~Lee, and A.~Saxena, ``{Deep Learning for Detecting Robotic
  Grasps},'' \emph{Intl. Journal of Robotics Research}, vol.~34, no. 4-5, pp.
  705--724, 2015.

\bibitem{pinto2016supersizing}
L.~Pinto and A.~Gupta, ``{Supersizing Self-supervision: Learning to Grasp from
  50K Tries and 700 Robot Hours},'' in \emph{IEEE Intl. Conf. on Robotics and
  Automation}, 2016, pp. 3406--3413.

\bibitem{Kopicki2016}
M.~Kopicki, R.~Detry, M.~Adjigble, R.~Stolkin, A.~Leonardis, and J.~L. Wyatt,
  ``{One-Shot Learning and Generation of Dexterous Grasps for Novel Objects},''
  \emph{Intl. Journal of Robotics Research}, vol.~35, no.~8, pp. 959--976,
  2016.

\bibitem{mahler2017dex}
J.~Mahler, J.~Liang, S.~Niyaz, M.~Laskey, R.~Doan, X.~Liu, J.~A. Ojea, and
  K.~Goldberg, ``{Dex-Net 2.0: Deep Learning to Plan Robust Grasps with
  Synthetic Point Clouds and Analytic Grasp Metrics},'' in \emph{Robotics:
  Science and Systems}, 2017.

\bibitem{lu2017grasp}
Q.~Lu, K.~Chenna, B.~Sundaralingam, and T.~Hermans, ``{Planning Multi-Fingered
  Grasps as Probabilistic Inference in a Learned Deep Network},'' in
  \emph{Intl. Symp. on Robotics Research}, 2017.

\bibitem{mousavian20196}
A.~Mousavian, C.~Eppner, and D.~Fox, ``6-dof graspnet: Variational grasp
  generation for object manipulation,'' \emph{arXiv preprint arXiv:1905.10520},
  2019.

\bibitem{sahbani2012overview}
A.~Sahbani, S.~El-Khoury, and P.~Bidaud, ``{An Overview of 3D Object Grasp
  Synthesis Algorithms},'' \emph{Robotics and Autonomous Systems}, vol.~60,
  no.~3, pp. 326--336, 2012.

\bibitem{varley2015generating}
J.~Varley, J.~Weisz, J.~Weiss, and P.~Allen, ``{Generating Multi-Fingered
  Robotic Grasps via Deep Learning},'' in \emph{IEEE/RSJ Intl. Conf. on
  Intelligent Robots and Systems}, 2015, pp. 4415--4420.

\bibitem{veres2017modeling}
M.~Veres, M.~Moussa, and G.~W. Taylor, ``{Modeling Grasp Motor Imagery through
  Deep Conditional Generative Models},'' \emph{{IEEE} Robotics \& Automation
  Letters}, vol.~2, no.~2, pp. 757--764, 2017.

\bibitem{lu2020multi}
Q.~Lu, M.~Van~der Merwe, B.~Sundaralingam, and T.~Hermans, ``Multi-fingered
  grasp planning via inference in deep neural networks,'' \emph{{IEEE} Robotics
  \& Automation Magazine}, 2020.

\bibitem{levine2016learning}
S.~Levine, P.~Pastor, A.~Krizhevsky, J.~Ibarz, and D.~Quillen, ``{Learning
  Hand-Eye Coordination for Robotic Grasping with Deep Learning and Large-Scale
  Data Collection},'' \emph{Intl. Journal of Robotics Research}, p.
  0278364917710318, 2016.

\bibitem{eppner2019billion}
C.~Eppner, A.~Mousavian, and D.~Fox, ``A billion ways to grasp: An evaluation
  of grasp sampling schemes on a dense, physics-based grasp data set,''
  \emph{arXiv preprint arXiv:1912.05604}, 2019.

\bibitem{settles2012active}
B.~Settles, \emph{{Active Learning}}.\hskip 1em plus 0.5em minus 0.4em\relax
  Morgan \& Claypool, 2012.

\bibitem{kroemer2010combining}
O.~Kroemer, R.~Detry, J.~Piater, and J.~Peters, ``Combining active learning and
  reactive control for robot grasping,'' \emph{Robotics and Autonomous
  Systems}, vol.~58, no.~9, pp. 1105--1116, 2010.

\bibitem{laskey2015multi}
M.~Laskey, J.~Mahler, Z.~McCarthy, F.~T. Pokorny, S.~Patil, J.~Van Den~Berg,
  D.~Kragic, P.~Abbeel, and K.~Goldberg, ``{Multi-armed bandit models for 2d
  grasp planning with uncertainty},'' in \emph{IEEE Intl. Conf. on Automation
  Science and Engineering}, 2015, pp. 572--579.

\bibitem{mahler2016dex}
J.~Mahler, F.~T. Pokorny, B.~Hou, M.~Roderick, M.~Laskey, M.~Aubry,
  K.~Kohlhoff, T.~Kr{\"o}ger, J.~Kuffner, and K.~Goldberg, ``{Dex-net 1.0: A
  cloud-based network of 3d objects for robust grasp planning using a
  multi-armed bandit model with correlated rewards},'' in \emph{IEEE Intl.
  Conf. on Robotics and Automation}, 2016, pp. 1957--1964.

\bibitem{lu2019grasp}
Q.~Lu and T.~Hermans, ``{Modeling Grasp Type Improves Learning-Based Grasp
  Planning},'' \emph{{IEEE} Robotics \& Automation Letters}, 2019.

\bibitem{gualtieri2016high}
M.~Gualtieri, A.~ten Pas, K.~Saenko, and R.~Platt, ``{High Precision Grasp Pose
  Detection in Dense Clutter},'' in \emph{IEEE/RSJ Intl. Conf. on Intelligent
  Robots and Systems}, 2016, pp. 598--605.

\bibitem{johns2016deep}
E.~Johns, S.~Leutenegger, and A.~J. Davison, ``{Deep Learning a Grasp Function
  for Grasping under Gripper Pose Uncertainty},'' in \emph{IEEE/RSJ Intl. Conf.
  on Intelligent Robots and Systems}, 2016, pp. 4461--4468.

\bibitem{redmon2015real}
J.~Redmon and A.~Angelova, ``{Real-Time Grasp Detection using Convolutional
  Neural Networks},'' in \emph{IEEE Intl. Conf. on Robotics and Automation},
  2015, pp. 1316--1322.

\bibitem{kumra2016robotic}
S.~Kumra and C.~Kanan, ``{Robotic Grasp Detection using Deep Convolutional
  Neural Networks},'' in \emph{IEEE/RSJ Intl. Conf. on Intelligent Robots and
  Systems}, 2017.

\bibitem{kalashnikov2018qt}
D.~Kalashnikov, A.~Irpan, P.~Pastor, J.~Ibarz, A.~Herzog, E.~Jang, D.~Quillen,
  E.~Holly, M.~Kalakrishnan, V.~Vanhoucke \emph{et~al.}, ``Qt-opt: Scalable
  deep reinforcement learning for vision-based robotic manipulation,''
  \emph{arXiv preprint arXiv:1806.10293}, 2018.

\bibitem{wu2020generative}
B.~Wu, I.~Akinola, A.~Gupta, F.~Xu, J.~Varley, D.~Watkins-Valls, and P.~K.
  Allen, ``Generative attention learning: a “general” framework for
  high-performance multi-fingered grasping in clutter,'' \emph{Autonomous
  Robots}, pp. 1--20, 2020.

\bibitem{ciocarlie2007dexterous}
M.~Ciocarlie, C.~Goldfeder, and P.~Allen, ``{Dimensionality Reduction for
  Hand-Independent Dexterous Robotic Grasping},'' in \emph{IEEE/RSJ Intl. Conf.
  on Intelligent Robots and Systems}, 2007, pp. 3270--3275.

\bibitem{osa2016experiments}
T.~Osa, J.~Peters, and G.~Neumann, ``{Experiments with hierarchical
  reinforcement learning of multiple grasping policies},'' in \emph{Int. Symp.
  on Exp. Robot.}\hskip 1em plus 0.5em minus 0.4em\relax Springer, 2016, pp.
  160--172.

\bibitem{morales2004active}
A.~Morales, E.~Chinellato, A.~H. Fagg, and A.~P. del Pobil, ``{An active
  learning approach for assessing robot grasp reliability},'' in \emph{IEEE/RSJ
  Intl. Conf. on Intelligent Robots and Systems}, 2004, pp. 485--490.

\bibitem{montesano2012active}
L.~Montesano and M.~Lopes, ``Active learning of visual descriptors for grasping
  using non-parametric smoothed beta distributions,'' \emph{Robotics and
  Autonomous Systems}, vol.~60, no.~3, pp. 452--462, 2012.

\bibitem{fu2019active}
X.~Fu, Y.~Liu, and Z.~Wang, ``Active learning-based grasp for accurate
  industrial manipulation,'' \emph{IEEE Transactions on Automation Science and
  Engineering}, 2019.

\bibitem{tiantransferring}
H.~Tian, C.~Wang, D.~Manocha, and X.~Zhang, ``{Transferring Grasp
  Configurations using Active Learning and Local Replanning},'' \emph{arxiv},
  2018.

\bibitem{laskey2015budgeted}
M.~Laskey, Z.~McCarthy, J.~Mahler, F.~T. Pokorny, S.~Patil, J.~Van Den~Berg,
  D.~Kragic, P.~Abbeel, and K.~Goldberg, ``{Budgeted multi-armed bandit models
  for sample-based grasp planning in the presence of uncertainty},'' in
  \emph{IEEE Intl. Conf. on Robotics and Automation}, 2015.

\bibitem{oberlin2018autonomously}
J.~Oberlin and S.~Tellex, ``{Autonomously acquiring instance-based object
  models from experience},'' in \emph{Intl. Symp. on Robotics Research}, 2015,
  pp. 73--90.

\bibitem{eppner2017visual}
C.~Eppner and O.~Brock, ``{Visual detection of opportunities to exploit contact
  in grasping using contextual multi-armed bandits},'' in \emph{IEEE/RSJ Intl.
  Conf. on Intelligent Robots and Systems}, 2017, pp. 273--278.

\bibitem{vandermerwe-icra2020-reconstruction-grasping}
\BIBentryALTinterwordspacing
M.~V. der Merwe, Q.~Lu, B.~Sundaralingam, M.~Matak, and T.~Hermans, ``{Learning
  Continuous 3D Reconstructions for Geometrically Aware Grasping},'' in
  \emph{IEEE International Conference on Robotics and Automation (ICRA)}, 2020.
  [Online]. Available:
  \url{https://sites.google.com/view/reconstruction-grasp/home}
\BIBentrySTDinterwordspacing

\bibitem{Sutton2018a}
R.~S. Sutton, A.~G. Barto, and A.~B. Book, \emph{{Reinforcement Learning : An
  Introduction}}.\hskip 1em plus 0.5em minus 0.4em\relax MIT Press, 2018.

\bibitem{gal2017deep}
Y.~Gal, R.~Islam, and Z.~Ghahramani, ``Deep bayesian active learning with image
  data,'' in \emph{Intl. Conf. on Machine Learning}, 2017.

\bibitem{singh2014bigbird}
A.~Singh, J.~Sha, K.~S. Narayan, T.~Achim, and P.~Abbeel, ``{Bigbird: A
  Large-Scale 3D Database of Object Instances},'' in \emph{IEEE Intl. Conf. on
  Robotics and Automation}, 2014, pp. 509--516.

\bibitem{jost2006entropy}
L.~Jost, ``Entropy and diversity,'' \emph{Oikos}, vol. 113, no.~2, pp.
  363--375, 2006.

\bibitem{spellerberg2003tribute}
I.~F. Spellerberg and P.~J. Fedor, ``A tribute to claude shannon (1916--2001)
  and a plea for more rigorous use of species richness, species diversity and
  the ‘shannon--wiener’index,'' \emph{Global ecology and biogeography},
  vol.~12, no.~3, pp. 177--179, 2003.

\bibitem{calli2015benchmarking}
B.~Calli, A.~Singh, A.~Walsman, S.~Srinivasa, P.~Abbeel, and A.~M. Dollar,
  ``{The YCB Object and Model Set: Towards Common Benchmarks for Manipulation
  Research},'' in \emph{Intl. Conf. on Advanced Robotics}, 2015, pp. 510--517.

\end{thebibliography}

\clearpage
\end{document}